\documentclass[letterpaper]{article} % DO NOT CHANGE THIS
\usepackage{aaai25}  % DO NOT CHANGE THIS
\usepackage{times}  % DO NOT CHANGE THIS
\usepackage{helvet}  % DO NOT CHANGE THIS
\usepackage{courier}  % DO NOT CHANGE THIS
\usepackage[hyphens]{url}  % DO NOT CHANGE THIS
\usepackage{graphicx} % DO NOT CHANGE THIS
\usepackage{natbib}  % DO NOT CHANGE THIS AND DO NOT ADD ANY OPTIONS TO IT
\usepackage{caption} % DO NOT CHANGE THIS AND DO NOT ADD ANY OPTIONS TO IT
\usepackage{algorithm}
\usepackage{algorithmic}

\usepackage{booktabs}
\usepackage{multirow}
\usepackage{amssymb}        % twoheadrightarrow
\usepackage{amsmath}       % more options for equation (split lines, underset)
\usepackage{subcaption}     % subtable2

\usepackage{array}
\usepackage{ragged2e}

\usepackage{newfloat}
\usepackage{listings}
\DeclareCaptionStyle{ruled}{labelfont=normalfont,labelsep=colon,strut=off} % DO NOT CHANGE THIS
\floatstyle{ruled}
\newfloat{listing}{tb}{lst}{}
\floatname{listing}{Listing}
\title{Zero-Shot Scene Change Detection}
\author{
    Kyusik Cho\textsuperscript{\rm 1}, 
    Dong Yeop Kim\textsuperscript{\rm 2,1}, 
    Euntai Kim\textsuperscript{\rm 1}\thanks{Corresponding author.}
}
\affiliations{
    \textsuperscript{\rm 1}Yonsei University, Seoul, Republic of Korea \\
    \textsuperscript{\rm 2}Korea Electronics Technology Institute, Seoul, Republic of Korea \\
    ks.cho@yonsei.ac.kr, sword32@keti.re.kr, etkim@yonsei.ac.kr
}

\usepackage{bibentry}
\begin{document}

\maketitle

\begin{abstract}
We present a novel, training-free approach to scene change detection.
Our method leverages tracking models, which inherently perform change detection between consecutive frames of video by identifying common objects and detecting new or missing objects. 
Specifically, our method takes advantage of the change detection effect of the tracking model by inputting reference and query images instead of consecutive frames.  
Furthermore, we focus on the content gap and style gap between two input images in change detection, and address both issues by proposing adaptive content threshold and style bridging layers, respectively. 
Finally, we extend our approach to video, leveraging rich temporal information to enhance the performance of scene change detection.
We compare our approach and baseline through various experiments. While existing train-based baseline tend to specialize only in the trained domain, our method shows consistent performance across various domains, proving the competitiveness of our approach. 
% The code is available at
% \mbox{\url{https://github.com/kyusik-cho/ZSSCD}}.

\end{abstract}

\begin{links}
    \link{Code}{https://github.com/kyusik-cho/ZSSCD}
\end{links}

\section{Introduction}
Scene Change Detection (SCD) is the task that aims to detect differences between two scenes separated by a temporal interval. Recently, SCD has gained significant interest in various applications involving mobile drones and robots. For instance, drone-based SCD has been studied for land terrain monitoring~\cite{lv2023iterative, song2019small, agarwal2019development}, construction progress monitoring~\cite{han2021change}, and urban feature monitoring~\cite{chen2016building}. Additionally, SCD using mobile robots has also been researched for natural disaster damage assessment~\cite{jst2015change,sakurada2013detecting}, urban landscape monitoring~\cite{alcantarilla2018street, sakurada2020weakly}, and industrial warehouse management~\cite{park2021changesim,park2022dual}.

Recently, SCD has been tackled using deep learning. Deep learning-based SCD techniques follow a procedure of learning from a training dataset and applying the model to a test dataset. These approaches tend to face two main challenges: dataset generation costs and susceptibility to style variations. Firstly, creating a training dataset for SCD models is labor-intensive and costly. Recent research has focused on reducing these costs through semi-supervised~\cite{lee2024semi, sun2022semisanet} and self-supervised learning~\cite{seo2023self, 9340840} methods, as well as the use of synthetic data~\cite{Sachdeva_WACV_2023, lee2024semi}. While these approaches mitigate the expense of labeling, they often overlook the cost of acquiring image pairs, which arises from the substantial temporal intervals to capture changes. Secondly, due to the substantial temporal intervals between pre-change and post-change images, variations in seasons, weather, and time introduce significant differences in their visual characteristics. Consequently, SCD techniques must be robust to these style variations to be effective. However, the training dataset cannot include all the style variations present in real-world scenarios, making the trained model vulnerable to style variations that are not included in the training set.

\begin{figure*}[t]
  \centering
  \includegraphics[width=\linewidth]{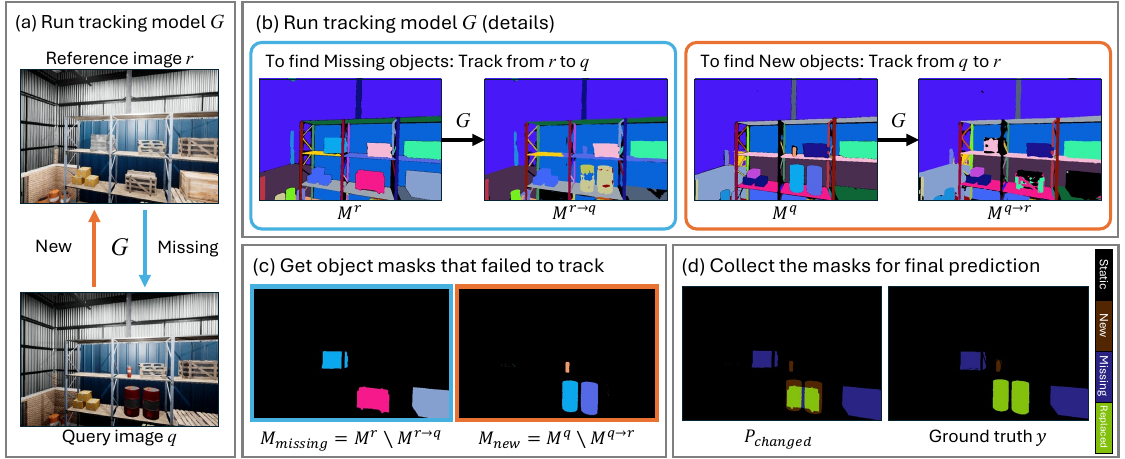}
  \caption{\textbf{The basic idea of SCD with tracking model.} (a) We execute the tracking model $G$ with $r$ and $q$. (b) We denote the tracking result from $r$ to $q$ as $M^{r \to q} = G(r, q, M^r)$, and the tracking result from $q$ to $r$ as $M^{q \to r} = G(q, r, M^q)$. (c) `Missing' objects are the objects that exist in $r$ but not in $q$. Therefore, we compare $M^r$ and $M^{r \to q}$ to find missing objects. Conversely, `new' objects are identified by comparing $M^q$ and $M^{q \to r}$. (d) The final prediction is the simple combination of new and missing.   
 }
  \label{figure_basic}
\end{figure*}

To address these problems, we propose a novel training-free zero-shot SCD method. Our method does not require a training dataset, thereby eliminating collection costs and allows it to be applied to any problem with arbitrary styles. To the best of our knowledge, this paper is the first to attempt zero-shot SCD without training on SCD datasets. 
The key idea of this paper is to formulate SCD as a tracking problem, and apply a foundation tracking model to conduct zero-shot SCD. This idea stems from the observation that the tracking task is fundamentally similar to change detection. Specifically, tracking models~\cite{cheng2023tracking, cheng2022xmem} maintain or build tracks by identifying the same objects, disappeared objects, and newly appeared objects in two consecutive images, even when the camera and objects move. Thus, if the two consecutive images in tracking are replaced with the two images before and after the change in SCD, the tracking model can automatically solve SCD without training.

However, there are some differences between tracking and SCD tasks:
(a) Unlike tracking, the two images before and after the change in SCD might have different styles due to a large time gap between two images. We refer to this SCD trait as the style gap. 
(b) Objects change very little between two consecutive images in tracking, whereas objects change abruptly in SCD. We refer to this SCD trait as the content gap.
To address these issues in our zero-shot SCD method, we introduce a style bridging layer and a content threshold, respectively.

Finally, we extend our approach to sequence, to introduce the zero-shot SCD approach works on the video. Since our approach operates based on a tracking model, it can be seamlessly extended to work with video sequences. The proposed zero-shot SCD approach has been evaluated on three benchmark datasets and demonstrated comparable or even superior performance compared to previous state-of-the-art training-based SCD methods.

\section{Related Work}
\subsubsection{Scene Change Detection (SCD)} 
In recent years, numerous deep learning-based change detection methods have been proposed for SCD.
DR-TANet~\cite{chen2021dr} utilized attention mechanism based on the encoder-decoder architecture.
SimSaC~\cite{park2022dual} developed a network with a warping module to correct distortions between images.
C-3PO~\cite{wang2023reduce} developed a network that fuses temporal features to distinguish three change types.
However, various studies have aimed to address the challenge of obtaining data. 
%For instance, \cite{lee2024semi, sun2022semisanet} have introduced semi-supervised learning, and \cite{seo2023self, 9340840} proposed the self-supervised learning with unlabeled data. However, \cite{Sachdeva_WACV_2023, lee2024semi} utilized synthetic data to effectively increase the dataset.
For instance, Lee and Kim \shortcite{lee2024semi} and Sun et al.~\shortcite{sun2022semisanet} have introduced semi-supervised learning, while Seo et al.~\shortcite{seo2023self} and Fukawa et al.~\shortcite{9340840} proposed the self-supervised learning with unlabeled data. However, Sachdeva and Zisserman~\shortcite{Sachdeva_WACV_2023} and Lee and Kim~\shortcite{lee2024semi} utilized synthetic data to effectively increase the dataset.
Despite these methods effectively reducing the label costs, they tend to overlook the cost of collecting image pairs. Moreover, the robustness against style change has not been previously discussed.
An effective SCD method should be able to focus on content changes regardless of variations in image style. However, since a single dataset cannot encompass all possible style variations, the performance tends to be specialized for the styles present in the dataset. This issue, while not evident in controlled laboratory environments, becomes a significant problem in real-world applications.
Therefore, we propose a novel SCD method that does not rely on datasets. Our method operates without a training dataset, thus ensuring independence from specific styles.

\subsubsection{Segmenting and Tracking Anything} 
Recently, Segment Anything~\cite{kirillov2023segment} has demonstrated highly effective performance in universal image segmentation. SAM has shown the ability to perform various zero-shot tasks, and has served as the foundational model for various studies~\cite{peng2023generalized, maquiling2024zero}. Building upon this research, researchers have explored various methods to extend its application to tracking.
For example, SAM-Track~\cite{cheng2023segment} implemented tracking by combining SAM with the DeAOT~\cite{yang2022decoupling} mask tracker. SAM-PT~\cite{rajivc2023segment} integrated SAM with point tracking to develop the pipeline. DEVA~\cite{cheng2023tracking} proposed a pipeline that uses the XMem tracker~\cite{cheng2022xmem} to track provided masks without additional training.
Among various studies, we adopted DEVA with SAM masks as our tracking model, to achieve track-anything for SCD without further training.

\section{Method}
Each datum for scene change detection (SCD) is represented as a triplet $(r, q, y)$, where $r$ and $q$ denote paired images acquired at distinct times $t_0$ and $t_1$, respectively, and $y$ represents the change label between the image pair.
The primary objective of this task is to discern the scene change between the images captured at $t_0$ and $t_1$ when inspecting the latter. Herein, we call the image obtained at $t_0$ ($r$) as the reference image and the image acquired at $t_1$ ($q$) as the query image.

To perform scene change detection without training, our methodology integrates two pretrained models: a segmentation model $F$ and a tracking model $G$. 
The segmentation model $F$ segments images in an unsupervised manner, while the tracking model $G$ tracks each mask generated by $F$ across multiple images. 
We employ the Segment Anything Model (SAM)~\cite{kirillov2023segment} as the segmentation model $F$ and DEVA~\cite{cheng2023tracking} for the tracking model $G$. Comprehensive details about parameters for $F$ and $G$, and details about the mask generation process are provided in the supplementary materials.

The rest of the Method section is structured as follows: First, we introduce the basic idea for performing SCD between two images using $F$ and $G$. 
Next, we discuss the differences between the tracking task and the SCD task, and then introduce methods to overcome these differences. 
Finally, we extend our approach to the video level.

\subsection{Scene Change Detection with Tracking Model}
\label{section_method_tracking2scd}
Our approach uses two pretrained models, a segmentation model $F$ and a tracking model $G$. 
The segmentation model $F$ partitions image $I$ into object-level masks, forming the set \mbox{$M = \{m_1, m_2, \cdots, m_n\}$}. There exists no overlap between distinct masks, that is, \mbox{$ m_i \cap m_j = 0, \forall~ i \neq j$}.
The tracking model $G$ takes consecutive frame images $(I^0, I^1)$ and the object masks of the first frame $M^0 = F(I^0)$ as input, and yields $M^{0 \to 1}$ as output, that is, \mbox{$M^{0 \to 1} = G(I^0, I^1, M^0)$}.
Here, $M^{0 \to 1}$ represents the set of masks tracked from \mbox{$M^0 = F(I^0)$ to $I^1$.} 
By checking if each mask that was present in $M^{0}$ is also present in $M^{0 \to 1}$, we can determine which object masks in $I^0$ still exist or have disappeared in $I^1$. 

The key idea of our zero-shot SCD approach is to apply a reference image $r$ and a query image $q$ instead of consecutive frames $(I^0, I^1)$ to the tracking model $G$. Although the tracking model traditionally expects consecutive frames $(I^0, I^1)$ for input, we deviate from this convention by providing reference image $r$ and query image $q$ instead. To avoid potential confusion, we rewrite the input and the output of the tracking model as $M^{r \to q} = G(r, q, M^r)$. By comparing the masks between $M^r$ and $M^{r \to q}$, we identify object masks that exist at time $t_0$ but have disappeared at time $t_1$, corresponding to the `missing' class in the change detection task. Specifically,
\begin{equation}
M_{missing} = M^{r} \setminus M^{r \to q}.
\label{eq_essence}
\end{equation}

Additionally, we run the tracking model $G$ again by reversing the order reference image $r$ and query image $q$ and feed them to $G$ to obtain $M^{q \to r} = G(q, r, M^q)$. Similarly, we predict the `new' objects by $M_{new} = M^{q} \setminus M^{q \to r}$, which represent the objects that appear at time $t_1$ but were absent at time $t_0$. 
Our pixel-wise prediction is obtained by applying the union of masks within $M_{new}$ and $M_{missing}$. Pixels experiencing both `new' and `missing' occurrences are considered `replaced.' Formally, change prediction $P_{changed}$ is determined by:
\begin{equation}
\begin{alignedat}{2}
   &P_{missing} &&= \bigcup M_{missing} \\
   &P_{new} &&= \bigcup M_{new}          \\
   &P_{replaced} &&= P_{missing} \cap P_{new} \\
   &P_{changed} &&= P_{missing} \cup P_{new}
\end{alignedat}
\end{equation}

The entire process of this approach is illustrated in Figure~\ref{figure_basic}.

\begin{figure}[t]
\includegraphics[width=0.48\textwidth]{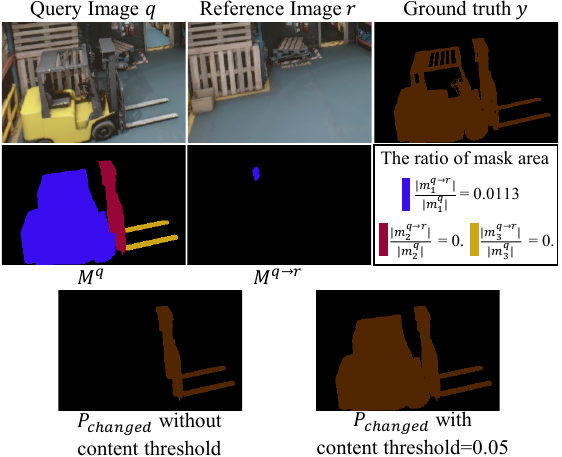}
    \caption{\textbf{Illustration of the content threshold.} 
    Since the yellow forklift in $q$ has disappeared in $r$, all the three masks (blue, red, and yellow masks) in $M^q$ have no associated masks in $M^{q \to r}$.  However, the tracking model creates a small area of the blue mask in $M^{q \to r}$ due to the content gap. This makes it mistakenly classified as a static object. To address this, we propose a content threshold to filter out masks whose area significantly reduces after tracking.
    }
\label{figure_CT}
\end{figure}

\subsection{Addressing Content Gap and Style Gap}
\label{section_method_2gaps}
As presented in the previous section, the key idea of our zero-shot SCD approach is to exploit the similarity between SCD and tracking tasks. However, directly applying this concept to various SCD scenarios leads to suboptimal performance due to inherent differences between the two tasks. In this section, we analyze these differences and propose corresponding solutions.

The first difference is the content gap, which refers to abrupt changes in content between the reference and query images. In traditional tracking tasks, objects typically disappear gradually over multiple frames rather than suddenly, and new objects appear gradually over multiple frames, implying that tracking tasks have little content gap.  
In contrast, SCD involves abrupt changes where objects disappear or appear within a single frame and it has a large content gap.
Therefore, when the tracking model $G$ trained on the tracking dataset is directly applied to SCD, the tracking model $G$ tends to create small segments even for objects that have disappeared, as shown in Figure~\ref{figure_CT}. In the first row, the yellow forklift in the query image is missing from the reference image, but, in the second row, the mask in $M^{q \to r}$  tracked from a blue mask in $M^q$ has a small segment. This remaining small segment makes the identification of missing objects very difficult. 

To address the content gap, we propose considering an object as disappeared if its size is significantly reduced after tracking, even if it has not completely vanished.
To the end, we introduce a content threshold $\tau$ and we compare the areas of the masks before and after tracking. If the ratio is less than the content threshold $\tau$, we consider the corresponding object is missing or newly appeared. 
We define the $\setminus\!^{\tau}$ operator to replace the $\setminus$ operator in equation~\ref{eq_essence} as follows, where $|m^*_i|$ denotes the area of the $i$-th mask in set *:
\begin{equation}
A \setminus\!\!^{\tau}\, B := \{m^A_i \, | \, { \frac{|m^B_i|}{\,|m^A_i|\,} } < \tau, \quad \forall m^A_i \in A \} .
\end{equation}
By introducing this operator, we have \mbox{$M_{missing} = M^{r} \setminus\!^{\tau}\, M^{r \to q}$}.
The value of $\tau$ is automatically determined based on the input data. Further details and discussions on the determination of $\tau$ are provided in the supplementary material.

The second difference between tracking tasks and SCD is the style gap, which refers to the difference in style between the reference and query images. Change detection data have large temporal gaps, therefore lighting, weather, or season can change. These changes are commonly modeled as style changes~\cite{tang2023improved}. We define such variations as the style gap, which is not considered in traditional tracking tasks and can thus significantly degrade SCD performance.

\begin{figure}[t]
\includegraphics[width=0.48\textwidth]{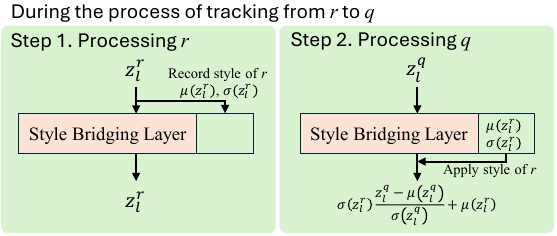}
\caption{\textbf{Illustration of the style bridging layer.} During the processing of the first image, the style is saved while the feature is passed through unchanged. When processing the second image, the saved style is applied to the feature.}
\label{figure_SBL}
\end{figure}
To address the style gap, we introduce a style bridging layer (SBL) by incorporating an Adaptive Instance Normalization (AdaIN) layer~\cite{huang2017arbitrary} into the residual blocks of ResNet backbone~\cite{he2016deep} of the tracking model $G$. 
The AdaIN layer, widely used to reduce style differences between images across various fields~\cite{karras2019style, wu2019ace, xu2023simde}, references the first image and applies its style to the second image, thereby reducing the style differences between the two images. Inspired by this, SBL addresses the style gap between two inputs without learning, with two training-free style parameters. 
For example, during the process \mbox{$M^{r \to q} = G(r, q, M^r)$}, the style bridging layer records the mean and variance of each layer in image $r$ and applies these statistics when processing image $q$. 
Formally, the style bridging layer updates the feature of $z^q_l$ as follows, where $z^*_l$ denotes the $l$-th layer feature of image *.
\begin{equation}
\tilde{z}^q_l  = \sigma(z^r_l) { \frac{ \, z^q_l - \mu(z^q_l) \, }{\sigma(z^q_l)} } +  \mu(z^r_l).
\end{equation}
The operation of the proposed style bridging layer is illustrated in Figure~\ref{figure_SBL}.

Through these two methods, we effectively and simply address the content gap and style gap. Note that the two improvements are also applied in the process \mbox{$M^{q \to r} = G(q, r, M^q)$} and \mbox{$M _{new} = M^{q} \; \setminus\!\!^{\tau} \; M^{q \to r}$}.

\subsection{Extension to the Video Sequences}
\label{section_method_tovideo}
In this subsection, we extend our image-based SCD approach to video sequences. Leveraging video data enhances spatial understanding by capturing scenes from multiple angles. 
The tracking model in our pipeline allows the seamless extension of our image-based SCD approach to video data.

\begin{figure*}[t]
  \centering
  \includegraphics[width=\linewidth]{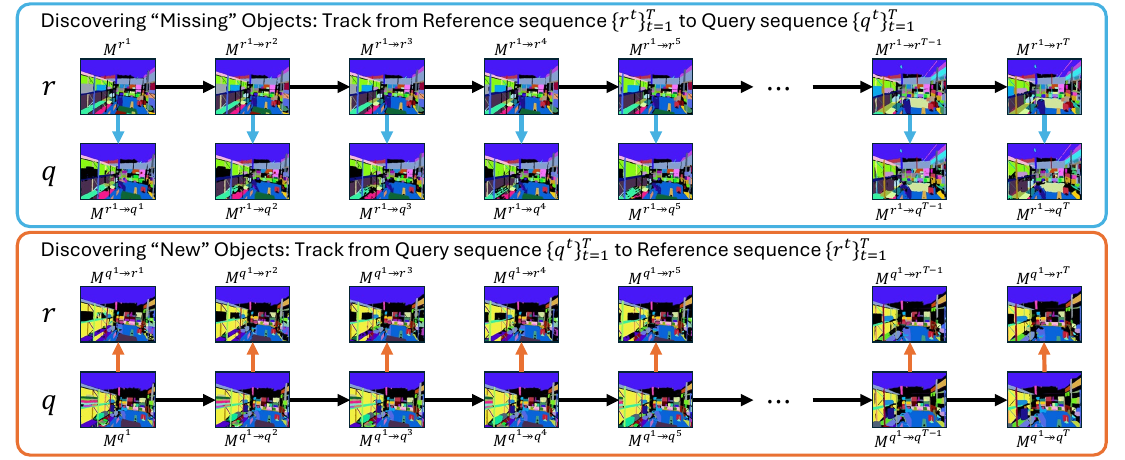}
  \caption{\textbf{Zero-shot SCD in video.} We conduct SCD on video sequences by providing sequence pairs instead of image pairs as input to the tracking model $G$. For each frame, the mask is propagated from the previous frame, resulting in a mask sequence through repeated propagation. SCD in the video is finalized by comparing the mask sequences. 
 }
  \label{figure_video}
\end{figure*}

Consider a video SCD dataset consisting of sequences of reference, query, and change labels, denoted as $\{r^t,q^t,y^t \}_{t=1}^T$, where $T$ represents the length of the video sequence, and $t$ denotes the time index. Compared to the image SCD, the video SCD requires two modifications. The first modification is simply to feed two sequences $\{r^t,q^t \}_{t=1}^T$ instead of image pair $\{r,q\}$ as input to the tracking model. Specifically, we start tracking to detect missing objects in the video with 
\begin{equation}
\begin{alignedat}{2}
  & M^{r^1 \to r^2} &&= G(r^1, r^2, M^{r^1}) \\[-3pt]
  & M^{r^1 \to r^2 \to q^2} &&= G(r^2, q^2, M^{r^1 \to r^2}) ~,
\end{alignedat}
\end{equation}
and continue tracking throughout the entire video with
\begin{equation}
\begin{alignedat}{2}
  & M^{r^1 \twoheadrightarrow r^t} &&= G(r^{t-1}, r^t, M^{r^1 \twoheadrightarrow r^{t-1}}) \\[-3pt]
  & M^{r^1 \twoheadrightarrow q^t} &&= G(r^t, q^t, M^{r^1 \twoheadrightarrow r^t}) ~ ,
\end{alignedat}
\end{equation}
where $M^{r^1 \twoheadrightarrow r^t} = M^{r^1 \to r^2 \to \cdots \to r^{t-1} \to r^t}$ and $M^{r^1 \twoheadrightarrow q^t} = M^{r^1 \to r^2 \to \cdots \to r^{t-1} \to r^t \to q^t }$, as shown in Figure~\ref{figure_video}. In mask formulation, $M^X$ denotes the output from the segmentation model $F$, whereas $M^{X \to Y}$ or $M^{X \twoheadrightarrow Y}$ denotes the output from tracking model $G$. 
This architecture is similar to the structure of the Bayes filters or Markov process~\cite{thrun2002probabilistic}, in that all the information processed from $1$ to $t-1$ are included in $M^{r^1 \twoheadrightarrow r^{t-1}}$. 
Consequently, $M^{r^1 \twoheadrightarrow r^t}$ can be incrementally updated from $M^{r^1 \twoheadrightarrow r^{t-1}}$ and ${r^t,q^t}$, without reprocessing all previous images. 
During the incremental update from $M^{r^1 \twoheadrightarrow r^{t-1}}$ to $M^{r^1 \twoheadrightarrow r^t}$, all functions for tracking, including updating feature memory and identifying new objects, are activated. Conversely, during the update from $M^{r^1 \twoheadrightarrow r^t}$ to $M^{r^1 \twoheadrightarrow q^t}$, these functions are deactivated. This processing sequence is designed to detect `missing' objects, whereas the opposite processing sequence with swapping $r$ and $q$ is employed to detect `new' objects.

The second modification is redefining $M^t_{missing}$ and $M^t_{new}$ to suit video data. 
In equation~\ref{eq_essence}, we defined $M_{missing}$ in image SCD as the masks that exist in the reference image $r$ but are absent in the query image $q$. 
The definition of $M^t_{missing}$ for video SCD is an extension of this definition to video. Specifically, $M^t_{missing}$ is defined as the masks present in the reference sequence $\{r^t \}_{t=1}^T$ but absent in the query sequence $\{q^t \}_{t=1}^T$.
Formally, $M^t_{missing}$ at time index $t$ is defined by:
\begin{equation}
\begin{alignedat}{2}
& M_{missing}^t = && \;\;  \{ M^{r^t} \setminus\!\!^{\tau}\, M^{r^1 \twoheadrightarrow q^1} \} \\[-3pt] 
& &&\cap \{ M^{r^t} \setminus\!\!^{\tau}\, M^{r^1 \twoheadrightarrow q^2} \} \\
& &&\cap \;\; \cdots \\[-3pt]
& &&\cap \{ M^{r^t} \setminus\!\!^{\tau}\, M^{r^1 \twoheadrightarrow q^T} \} .
\end{alignedat}
\end{equation}
$M^t_{new}$ is similarly defined with the object that exists in query sequence $\{q^t \}_{t=1}^T$, but is absent in the reference sequence $\{r^t \}_{t=1}^T$. 

Through these two extensions, our methodology becomes appropriate for processing videos.
According to the definitions of these modifications, the sequence length can range from 1 to infinity. 
However, we impose an upper bound on the length of a sequence, denoted as $T_{max}$. If the length of the video exceeds $T_{max}$, it is divided into multiple sequences, each with a length of $T_{max}$. 
The reason for constraining the length of sequences is simple: as sequences lengthen, memory costs increase, while the relevant information for change detection decreases. For instance, in scenarios where the camera is in motion, the initial and final frames of a sequence may capture entirely different locations, rendering them unsuitable for change detection. Conversely, if the camera remains stationary, all frames depict the same scene, and additional frames provide redundant information. Therefore, we ensure more effective SCD by setting the upper bound of the sequence length. 
For our experiments, we set $T_{max}$ to 60.

\begin{table*}[t]
\centering
\setlength{\tabcolsep}{7.5pt}
\begin{tabular}{ccc|ccccc}
\toprule
\multicolumn{8}{c}{ChangeSim: In-domain} \\
\midrule
{~~~~~~Method~~~~~~}  & ~~Trained Set~~ & ~~~~Test Set~~~~ & ~~Static~ & ~~New~~ & Missing & Replaced & ~~~mIoU~~~                         \\
\midrule
C-3PO & Normal & \multirow{2}{*}{Normal}   & 94.2 & 14.3 & ~~5.3 & 17.1 & 32.7             \\
Ours                                  & -  & & 93.9 & 29.6 & 12.3 & ~~7.3 & \textbf{35.8}    \\
\midrule
C-3PO                       & Dusty-air &  \multirow{2}{*}{Dusty-air} & 94.0 & ~~9.3 & ~~2.8 & 12.6 & 29.7           \\
Ours                                  & -  & & 88.6 & 23.2 & ~~6.4 & ~~8.1 & \textbf{31.6}   \\
\midrule
C-3PO                       & ~Low-illum.~ &  \multirow{2}{*}{~~Low-illum.~~} & 93.8 & ~~5.4 & ~~0.6 & ~~8.4 & \textbf{27.1}  \\
Ours                                  & -  & & 80.6 & ~~9.4 & ~~4.7 & ~~6.3 & 25.2          \\
\bottomrule
\end{tabular}
\caption{\textbf{Experimental results on ChangeSim.}
The results are expressed in per-class IoU and mIoU scores.
Despite the absence of a training process, our model outperformed the baseline's in-domain performance in two out of three subsets.
}
\label{table_changesim}
\end{table*}

\begin{table}[t]
\centering
\setlength{\tabcolsep}{4.5pt}
\begin{tabular}{cc|ccc}
\toprule
\multicolumn{5}{c}{ChangeSim: Cross-domain} \\
\midrule
 & & \multicolumn{3}{c}{Test set}                        \\
{~Method~}  & Trained Set & Normal & Dusty-air & Low-illum.                      \\
\midrule
\multirow{3}{*}{C-3PO} & Normal &   32.7 & 27.2 & 26.7         \\
                         & Dusty-air &   29.6 & 29.7 & 26.9        \\
                         & Low-illum. &    29.4 & 27.1 & \textbf{27.1}          \\
\midrule
Ours                         & -         &  \textbf{35.8} & \textbf{31.6} & 25.2  \\
\bottomrule
\end{tabular}
\caption{\textbf{Experimental results on ChangeSim, cross-domain.}
The results are expressed in the mIoU score across all change classes. 
We trained the baseline model on each subset and tested it across all subsets. The experimental results show that the baseline model achieves the highest performance when the training set and test set are the same, while performance degrades when the training and test sets differ. In contrast, our method is free from this issue.
}
\label{table_changesim_cross_domain}
\end{table}

\begin{table}[ht]
\centering
\setlength{\tabcolsep}{2.5pt}
\begin{tabular}{cc|ccc}
\toprule
\multicolumn{5}{c}{VL-CMU-CD \& PCD} \\
\midrule
 & & \multicolumn{3}{c}{Test set}                        \\
{Method}  & Trained Set & VL-CMU-CD & ~~PCD~~ & ~Average~             \\
\midrule
\multirow{2}{*}{C-3PO} & VL-CMU-CD &   \textbf{79.4} & 11.6 & 45.5         \\
                         & PCD &    24.3 & \textbf{82.4} & 53.4          \\
\midrule
Ours                         & -         &  51.6 & 56.5 & \textbf{54.0} \\
\bottomrule
\end{tabular}
\vspace{0.2cm}
\caption{\textbf{Experimental results on VL-CMU-CD and PCD.}
The results are expressed in the F1 score. The baseline model performs best when the training and test are identical. However, its performance significantly declines when these datasets differ. Conversely, our method is robust to changes in the dataset, maintaining performance without the need for retraining whenever the test environment changes.
}
\label{table_VLCMU_PCD}
\end{table}

\section{Experiments}
\subsection{Experimental Setup}
\label{section_setup}
In this section, we briefly introduce the datasets, the relevant settings, and the evaluation metrics. 

% \subsubsection{ChangeSim}
\noindent\textbf{ChangeSim}~\cite{park2021changesim} is a synthetic dataset with an industrial indoor environment. It includes three subsets with varying environmental conditions: normal, low-illumination, and dusty air. The dataset categorizes changes into four types: new, missing, rotated, and replaced. 
Despite its variety of environmental variations and change classes, most baseline experiments on this dataset have evaluated only the binary change/unchange classification and have predominantly focused on the normal subset, leaving the dataset's full potential underexplored.
Therefore, we chose the state-of-the-art method, C-3PO~\cite{wang2023reduce}, and reproduced the results under the following conditions: using the original image size ($640\times480$) to fully utilize the rich information; and including all three subsets. Among the four change classes in this dataset, the rotated class, unlike others, involves slight angular changes of the same object rather than complete appearances or disappearances. We considered this as the object remaining static and integrated this class into static for evaluation.

% \subsubsection{VL-CMU-CD}
\noindent
\textbf{VL-CMU-CD}~\cite{alcantarilla2018street} is a dataset that includes information on urban street view changes over a long period, encompassing seasonal variations. Following the baseline approach, we performed predictions using $512\times512$-sized images. As the change class in this dataset is limited to a binary classification of the `missing' class, we used only the `missing' class for our three types of prediction.

% \subsubsection{PCD}
\noindent
\textbf{PCD}~\cite{jst2015change} is a dataset consisting of panoramic images and includes two subsets: GSV and TSUNAMI. Following the baselines, we performed predictions on reshaped images of size $256\times1024$. Each data point is classified into binary change or unchanged categories. We consider the detected new, missing, and replaced predictions into a `changed' class for evaluation.

\subsubsection{Evaluation Metrics}
Following the previous work, we employ the mean Intersection over Union (mIoU) metric for ChangeSim and F1 score for VL-CMU-CD and PCD datasets.

\begin{figure*}[t]
  \centering
  \includegraphics[width=\linewidth]{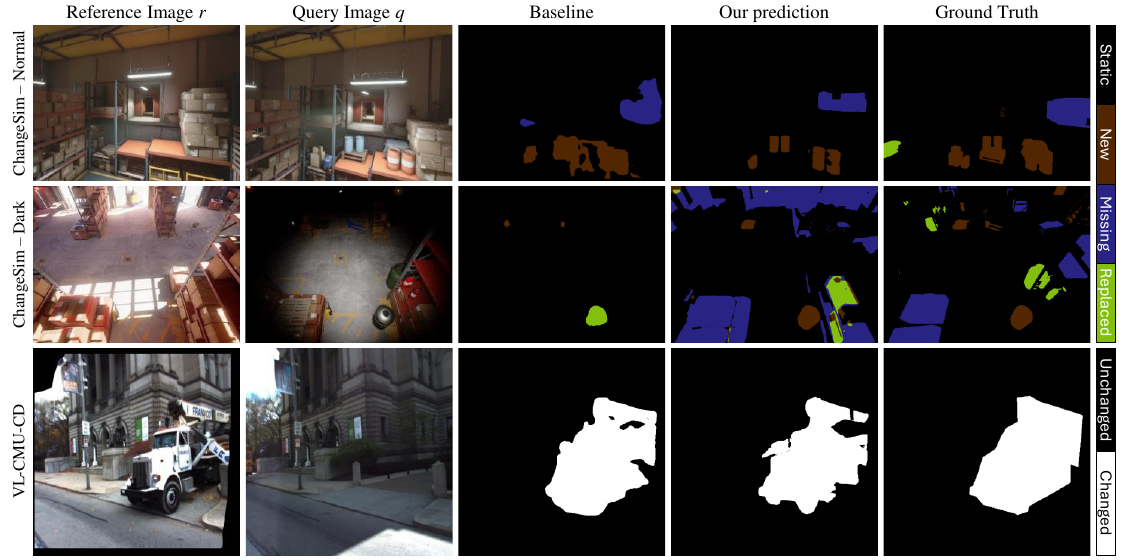}
  \caption{\textbf{Qualitative results.} Our approach successfully performs change detection across various datasets without training. For more qualitative results, see the supplementary material.
 }
  \label{figure_qualitative}
\end{figure*}

\subsection{Experimental Results}

Table~\ref{table_changesim} presents the experimental results with other state-of-the-art, C-3PO~\cite{wang2023reduce} from ChangeSim. 
C-3PO is reproduced under the conditions described in the previous section.
The baseline model is tested on the same datasets as its training dataset, denoted as in-domain.
The table shows that our model achieved superior performance in two out of three subsets: normal and dusty-air, with mIoU of 35.8 and 31.6, respectively. However, our model shows lower performance in low-illumination subset, with mIoU of 25.2. 

However, traditional train-based approaches are specialized for the style variation on which they were trained, becoming highly vulnerable when the domains differ.
To illustrate this, we conducted additional experiments testing the baseline model on different domains from the ones it was trained on. 
Specifically, we tested the baseline model trained on a particular subset of ChangeSim on the other subsets, denoted as cross-domain. The experimental results are shown in Table~\ref{table_changesim_cross_domain}.
The baseline model, being specialized for their in-domain data, suffers performance drops when the data changes, indicating a lack of generalization. 
Specifically, when the baseline model trained on the dusty-air subset is tested on the same subset, it achieves a mIoU of 29.7. However, when the model is trained on the normal or low-illumination subsets and tested on the dusty-air subset, the mIoU scores are relatively lower, at 27.2 and 27.1, respectively. This performance drop is also observed in other cross-domain experiments.
Conversely, our approach, which is not tailored to any specific style variation within the training dataset, shows robustness to changing the datasets.
In other words, our model can be applied to all subsets without the need for retraining each time the environment changes.

We also conducted experiments on two additional real-world datasets, VL-CMU-CD and PCD. The results are summarized in Table~\ref{table_VLCMU_PCD}. These results reveal a pattern consistent with our previous observations: the baseline model performs well when the training and test datasets are identical, but its performance significantly declines when the datasets differ.
In this experiment, the performance drop is particularly severe due to the greater stylistic differences between the datasets. Specifically, the baseline model achieves high in-domain performance with F1 scores of 79.4 and 82.4 on the VL-CMU-CD and PCD datasets, respectively. In contrast, its cross-domain performance drops dramatically to 24.3 and 11.6, respectively.
This observation suggests that the baseline model is overfitted to the limited style variations present in the training dataset. 

Since the change detection method must function across various seasons and weather conditions in real-world scenarios, robustness to style changes is crucial for reliable performance. 
The most straightforward way to achieve such robustness against style variations is to employ an approach that does not rely on the training set, as supported by these experimental results.

We present qualitative results in Figure~\ref{figure_qualitative}.
The baseline results correspond to in-domain scenarios, where the training and test datasets are identical. In contrast, our predictions are zero-shot results without any training. The qualitative results show that our approach effectively performs change detection without training. 
In the supplementary material, we provide more qualitative results and intermediate outcomes of the prediction process.

\subsection{Ablation Experiments}
We conducted extensive ablation experiments and more analyses in the supplementary materials to validate the efficacy and robustness of our proposed methods. (1) Performance of zero-shot SCD with and without the proposed adaptive content threshold (ACT) and style bridging layer (SBL). (2) Ablation study based on different SBL settings. (3) Ablation study based on different clip lengths. (4) Ablation study based on different ACT settings. %(5) More qualitative results. % and intermediate outcomes of the prediction process.

\section{Conclusion}
\label{sec:conclusion}
In this paper, we present a novel approach to zero-shot Scene Change Detection (SCD) applicable to both image and video. Our method performs SCD without training by leveraging a tracking model, which inherently performs change detection between consecutive video frames by recognizing common objects and identifying new or missing objects. To adapt the tracking model specifically for SCD, we propose two training-free components: the style bridging layer and the adaptive content threshold. Through extensive experiments on three SCD datasets, our approach demonstrates its versatility by showing its robustness to various environmental changes. We believe that our work offers a fresh perspective on SCD and represents a significant step forward in its practical application.

\section{Acknowledgments}
This work was supported by Institute of Information \& communications Technology Planning \& Evaluation (IITP) grant funded by the Korea government(MSIT) (No.2022-0-01025, Development of core technology for mobile manipulator for 5G edge-based transportation and manipulation)

% \clearpage
\bibliography{aaai25}
\clearpage

\twocolumn[
\vbox{%
  \hsize\textwidth%
  \linewidth\hsize%  
  \vskip 0.625in minus 0.125in%
  \centering%
  {\LARGE Supplementary Material of \bf ``Zero-Shot Scene Change Detection'' \par}%
    \vskip 0.4in
    \Large
    Technical Appendix \\
    \vskip 0.9in
}%
]

      % \vskip 0.1in plus 0.5fil minus 0.05in%
      % {\Large{\textbf{\theauthors\ifhmode\\\fi}}}%
      % \vskip .2em plus 0.25fil%
      % {\normalsize \affiliations_\ifhmode\\\fi}%
      % \vskip .5em plus 2fil%

\appendix

\section{Implementation Details}
\subsection{Details on Mask Generation}

This section explains how to create a mask for input into the tracking model.
We use the Segment Anything Model (SAM)~\cite{kirillov2023segment} to generate the mask proposal. We executed SAM's automatic mask generation pipeline with the parameters set as shown in Table~\ref{table_sam_param}.

\begin{table}[ht]
\centering
\setlength{\tabcolsep}{20pt}
\begin{tabular}{c|c}
\toprule
Hyperparameter & Value\\
\midrule
SAM model                                 & vit\_h  \\
points\_per\_side &32 \\ 
pred\_iou\_thresh &0.86 \\ 
stability\_score\_thresh &0.92 \\ 
crop\_n\_layers &1 \\ 
crop\_n\_points\_downscale\_factor &2 \\ 
min\_mask\_region\_area &100 \\
\bottomrule
\end{tabular}
% \vspace{0.1cm}
\caption{Hyperparameters of SAM.}
\label{table_sam_param}
\end{table}

Meanwhile, the masks generated by SAM exhibit two characteristics that make them unsuitable for our task. First, there are too many small masks. Second, a single pixel can be assigned to multiple masks. 
These characteristics are problematic because scene change detection (SCD) operates at the object level. Specifically, since the changes occur at the object level, the mask size should correspond to the object level and not be too small. Moreover, each pixel in the image should belong to only one object, meaning it must belong to at most one mask.
Therefore, we conducted the post-processing steps outlined in Table~\ref{table_mask_gen}. This process ensures that each pixel belongs to at most one mask, and small masks are naturally removed.

\begin{table}[!ht]
\centering
\setlength{\tabcolsep}{5pt}
\begin{tabular}{c|>{\justifying\arraybackslash}p{7cm}}
\toprule
Step & \noindent Details\\
\midrule
1 & \noindent Run the Segment Anything Model (SAM) to obtain masks. \\
2 & \noindent Sort the SAM masks from smallest to largest area. \\
3 & \noindent Overlay the sorted masks in order. This process ensures that if one pixel belongs to multiple masks, the largest mask among them is selected. \\
4 & \noindent For each mask, if the ratio of the area covered by the overlaid mask is more than 50\%, it is merged with the largest mask among the overlaid masks. This process removes the masks that occluded most of the area through step 3. \\
5 & \noindent For the VL-CMU-CD~\cite{alcantarilla2018street} dataset, remove the masks that do not contain any information. These are the black areas at the edges of the images. \\
\bottomrule
\end{tabular}
% \vspace{0.1cm}
\caption{Mask generation process.}
\label{table_mask_gen}
\end{table}

\subsection{Details on Tracking Model}
We use DEVA~\cite{cheng2023tracking} for our tracking model.
We used the DEVA structure with only one modification, incorporating style bridging layers (SBL) within the encoder architecture. The first SBL is positioned immediately after the first convolutional layer, while subsequent SBLs are placed after the addition operation within each residual block~\cite{he2016deep}. DEVA parameters were set as shown in Table~\ref{table_deva_param}.

Notably, the extension to video with $G$ reduces the need to run the segmentation model $F$ for every frame and reduces total computation cost.
This is based on the observation that most tracking models exploit the similarity between adjacent frames in a video to reduce computation: instead of performing computationally intensive operations on all frames, these operations are restricted to keyframes, while intermediate frames are processed through lightweight feature propagation~\cite{zhu2017deep, jain2019accel}.

In our experiments, $F$ is executed every 5 frames to get the set of object masks (denoted as detection\_every in Table~\ref{table_deva_param}), while other frames are propagated by $G$ from the previous frame.

\begin{table}[ht]
\centering
\setlength{\tabcolsep}{20pt}
\begin{tabular}{c|c}
\toprule
Hyperparameter & Value\\
\midrule
detection\_every & 5 \\
voting\_frames & 3 \\
max\_missed\_detection\_count & 5 \\
max\_num\_objects & 200 \\
\bottomrule
\end{tabular}
% \vspace{0.1cm}
\caption{Hyperparameters of DEVA.}
\label{table_deva_param}
\end{table}

\subsection{Details on Adaptive Content Threshold}

The set of missing object masks $M^t_{missing}$ defined by equation 7 in the main paper, are the masks that exist in the reference frame but are missing in the query sequence. If an object mask is detected in any frame of the query sequence, it is not classified as missing. Notably, even if an object mask is absent in the majority of frames within the query sequence, the detection of the object mask in just a single frame prevents it from being classified as missing. As a result, as the number of sequences being compared increases, the model becomes more vulnerable to noise, thereby complicating the accurate prediction of missing objects.

Therefore, we hypothesized that the content threshold should adapt to the length of the video sequence.
For the reasons mentioned above, a high content threshold risks missing too many true positives in short sequences, while a low content threshold becomes vulnerable to noise in long sequences.
To resolve this trade-off, we designed the adaptive content threshold as a parameter explicitly dependent on the video sequence length.

Specifically, we designed the adaptive content threshold to satisfy the following conditions: (1) It should increase as the clip length gets longer. (2) It should have a lower bound to ensure functionality at the image level. (3) It should have an upper bound to prevent excessive growth. 
Based on these considerations, we derived a simple equation for the adaptive content threshold that depends solely on the sequence length as follows: 
\begin{equation}
    \tau = 0.5 - {\frac{0.9}{\sqrt{length}+1}}.
\label{eq_adaptive_th}
\end{equation}

The effectiveness of this adaptive content threshold is further examined in our ablation studies.

\subsection{Computing Infrastructure}
The experiments were conducted using the Intel Xeon Gold 6426Y CPU and a single NVIDIA RTX A5000 GPU. The software stack employed includes PyTorch 2.1.2 with CUDA version 12.1. The size of memory usage is below 24 GB.

\section{Ablation Experiments}
We conducted ablation experiments to show the effectiveness of our approach. All experiments were conducted on the ChangeSim dataset~\cite{park2021changesim}.
To conserve space in the tables, the names of each subset are abbreviated: dusty-air is abbreviated as `Dust', and low-illumination as `Dark'.

\subsection{Addressing Content Gap and Style Gap} 
We evaluate the effectiveness of the proposed adaptive content threshold (ACT) and style bridging layer (SBL). As shown in Table~\ref{table_2gap}, experimental results indicate that the combined use of ACT and SBL yields the highest average performance. Additionally, these experiments offer interesting observations: (1) SBL is effective when the style of reference and query image differ (i.e., dusty-air and low-illumination subset), but its effectiveness diminishes in subsets with consistent styles (i.e., normal subset). (2) ACT demonstrates its efficacy particularly when the model's tracking performance is high (e.g., normal subset). In experiments where tracking performance is poor (e.g., low-illumination subset), the addition of ACT leads to a decline in performance.

\begin{table}[ht]
\centering
\begin{tabular}{cc|cccc}
\toprule
\multicolumn{2}{c|}{Config.} & \multicolumn{4}{c}{ChangeSim}                                                 \\
ACT         & SBL        & Normal                 & Dust                 & Dark                 & Average  \\
\midrule
                        &                 & 28.2  & 24.4  & 25.0  & 25.9 \\
                         &  \checkmark                & 27.1  & 27.0  & \textbf{25.7}  & 26.6 \\
\checkmark               &               & \textbf{36.0}  & 18.1  & 21.6  & 25.3 \\
\checkmark             & \checkmark                & 35.8  & \textbf{31.6}  & 25.2  & \textbf{30.9} \\

              \bottomrule
\end{tabular}
% \vspace{0.1cm}
\caption{ Ablation study on ACT and SBL. }
\label{table_2gap}
\end{table}

\subsection{The Number of SBL}
We experiment to determine the optimal number of SBL required. We progressively add SBL from the early layer of the encoder. The first Style Bridging Layer (SBL) was placed directly after the first convolutional layer, and the following SBLs were inserted after the addition operation within each residual block. The experimental results are shown in Table~\ref{table_sbl_number}. The experimental results indicate that applying SBL to all encoder blocks yields the best performance.

\begin{table}[ht]
\centering
\begin{tabular}{c|cccc}
\toprule
 & \multicolumn{4}{c}{ChangeSim} \\
~\# of SBL~  & Normal & ~Dust~ & ~Dark~ & Average  \\
\midrule
0  & 36.0  & 18.1  & 21.6  & 25.3  \\
1  & 35.7 & 29.2 & 23.9 & 29.6  \\
2  & \textbf{36.1} & 31.1 & 24.7 & 30.6 \\
3  & 36.0 & 31.5 & 25.0 & 30.8 \\ 
4  & 35.8 & \textbf{31.6} & \textbf{25.2} & \textbf{30.9} \\
\bottomrule
\end{tabular}
% \vspace{0.1cm}
\caption{ Ablation study on the number of SBL. }
\label{table_sbl_number}
\end{table}

\subsection{The Length of Sequence} 
We conduct experiments under various $T_{max}$ values. As shown in Table~\ref{table_seq_len}, our method shows a significant improvement when extended to video compared to the image-based SCD approach ($T_{max}=1$). However, it is notable that increasing $T_{max}$ does not consistently lead to improved performance; increasing the video length beyond 60 has little to no impact on performance or may even lead to a decline.

\begin{table}[ht]
\centering
\begin{tabular}{c|cccc}
\toprule
 & \multicolumn{4}{c}{ChangeSim} \\
$T_{max}$                                 & Normal             & ~Dust~             & ~Dark~             & Average  \\
\midrule
1 & 33.6 & 28.9 & 22.2 & 28.2 \\
30 & 35.6 & \textbf{31.6} & 24.4 & 30.5 \\
60 & \textbf{35.8} & 31.6 & 25.2 & \textbf{30.9} \\
90 & 35.6 & 31.2 &
\textbf{25.6} & 30.8 \\
unlimited & 33.3 & 30.5 & 25.6 & 29.8 \\
\bottomrule
\end{tabular}
% \vspace{0.1cm}
\caption{ Ablation study on the length of the sequence $T_{max}$. }
\label{table_seq_len}
\end{table}

\subsection{The Adaptive Content Threshold}
To illustrate the necessity of varying the content threshold according to the sequence length, we conducted experiments across three different sequence lengths.
The three sequence lengths were 1, 60, and 30, which are the $T_{max}$ for image-level SCD, our standard $T_{max}$ for video, and the intermediate value, respectively. The fixed threshold values were set to 0.05 and 0.4, approximating the values of ACT when $T_{max}=1$ and $T_{max}=60$, respectively.

As shown in Table~\ref{table_ACT}, when the sequence length is 1, a threshold of 0.05 performs the best, while performance is poor at a threshold of 0.4. Conversely, for sequence lengths of 30 and 60, a threshold of 0.05 results in the lowest performance, while higher thresholds improve performance. Furthermore, the results indicate that the ACT consistently achieves the best performance across all sequence lengths. This shows the validity and effectiveness of ACT and supports the argument that the threshold should be influenced by the sequence length.

\begin{table}[ht]
\centering
\begin{tabular}{c|c|cccc}
\toprule
 & \multicolumn{5}{c}{ChangeSim} \\
$T_{max}$                                 & $\tau$  & Normal             & Dust             & Dark             & Average  \\
\midrule
\multirow{3}{*}{1} & 0.05 & 33.6 &	28.9 &	22.2 &	28.2 \\
 & 0.4 &  31.9 & 	24.6 & 	18.4 & 	25.0 \\
 & Adaptive  & \textbf{33.6}	 & \textbf{28.9}	& \textbf{22.2}	& \textbf{28.2} \\
 \midrule
\multirow{3}{*}{30} & 0.05 &  31.1 &	29.9 &	\textbf{26.2} &	29.1 \\
 & 0.4 & 35.4 &	31.3 &	23.9 &	30.2 \\
 & Adaptive & \textbf{35.6} &	\textbf{31.6} &	24.4 &	\textbf{30.5} \\
 \midrule
\multirow{3}{*}{60} & 0.05 & 30.0 &	29.0 &	\textbf{26.5} &	28.5 \\
 & 0.4 & 35.8	& 31.5	& 25.2	& 30.8 \\
 & Adaptive & \textbf{35.8} & 	\textbf{31.6} & 	25.2 & 	\textbf{30.9} \\
\bottomrule
\end{tabular}
% \vspace{0.1cm}
\caption{ Ablation study on the content threshold ($\tau$). }
\label{table_ACT}
\end{table}

\section{Qualitative Results}
\begin{figure*}[t]
  \centering
  \includegraphics[width=\linewidth]{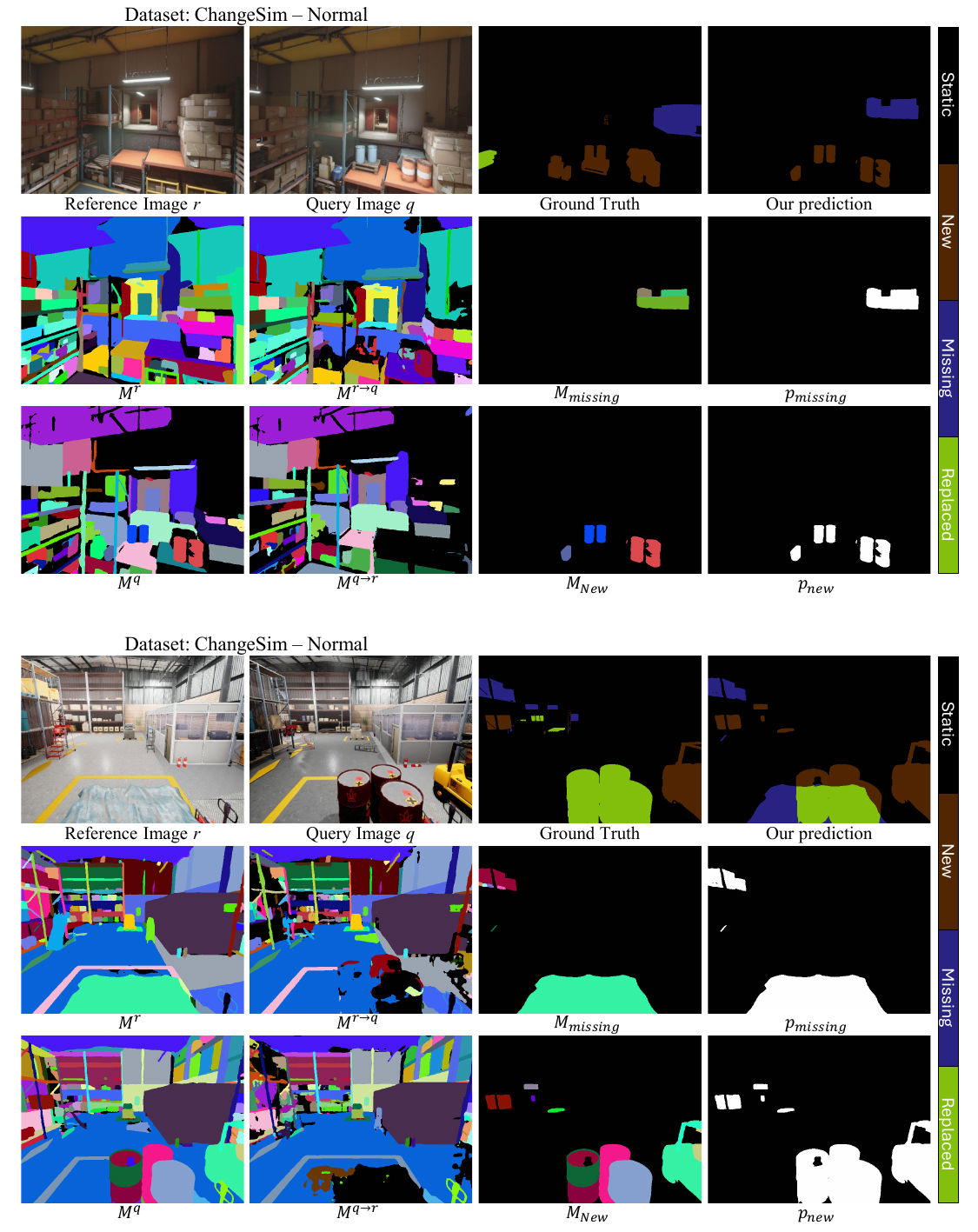}
  \vspace{-0.5cm}
  \caption{\textbf{Qualitative results on ChangeSim Normal subset.} 
 }
  \vspace{-0.2cm}
  \label{figure_qualitative1}
\end{figure*}

\begin{figure*}[t]
  \centering
  \includegraphics[width=\linewidth]{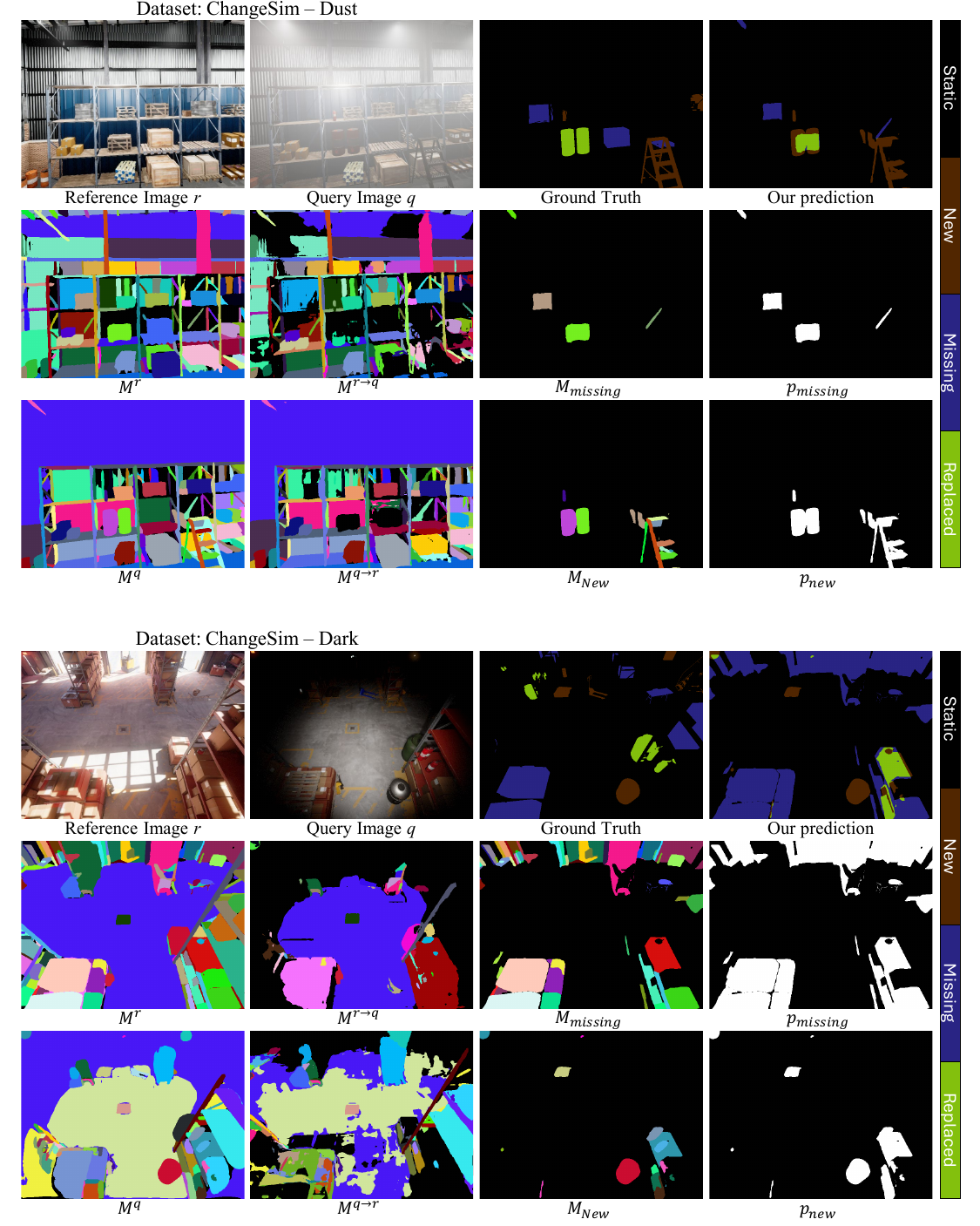}
  \vspace{-0.5cm}
  \caption{\textbf{Qualitative results on ChangeSim Low-illumination and ChangeSim Dusty-air.} 
 }
  \vspace{-0.2cm}
  \label{figure_qualitative2}
\end{figure*}

\begin{figure*}[t]
  \centering
  \includegraphics[width=\linewidth]{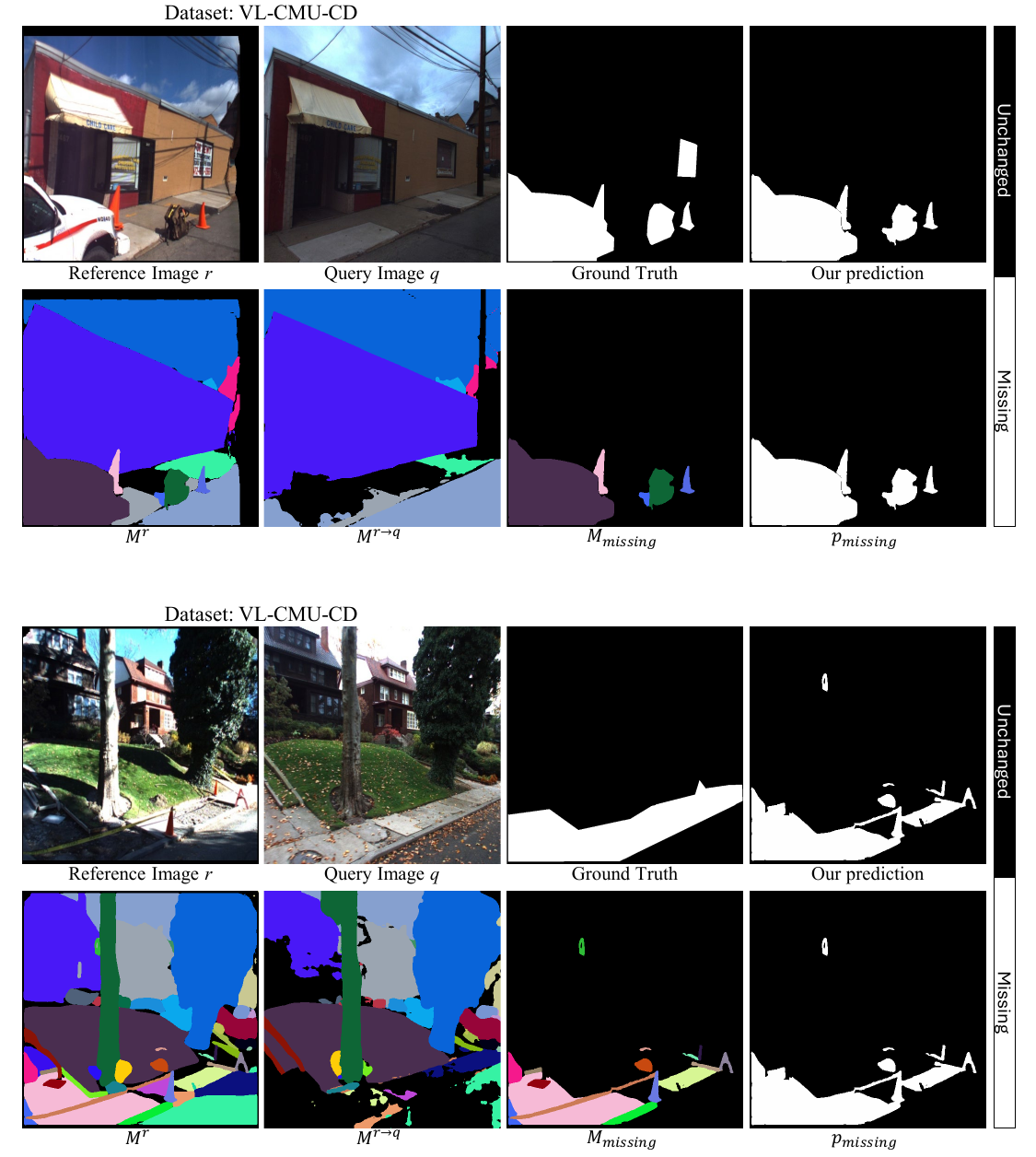}
  \vspace{-0.5cm}
  \caption{\textbf{Qualitative results on VL-CMU-CD~\cite{alcantarilla2018street}.} 
 }
  \vspace{-0.2cm}
  \label{figure_qualitative3}
\end{figure*}

\begin{figure*}[t]
  \centering
  \includegraphics[width=0.94\linewidth]{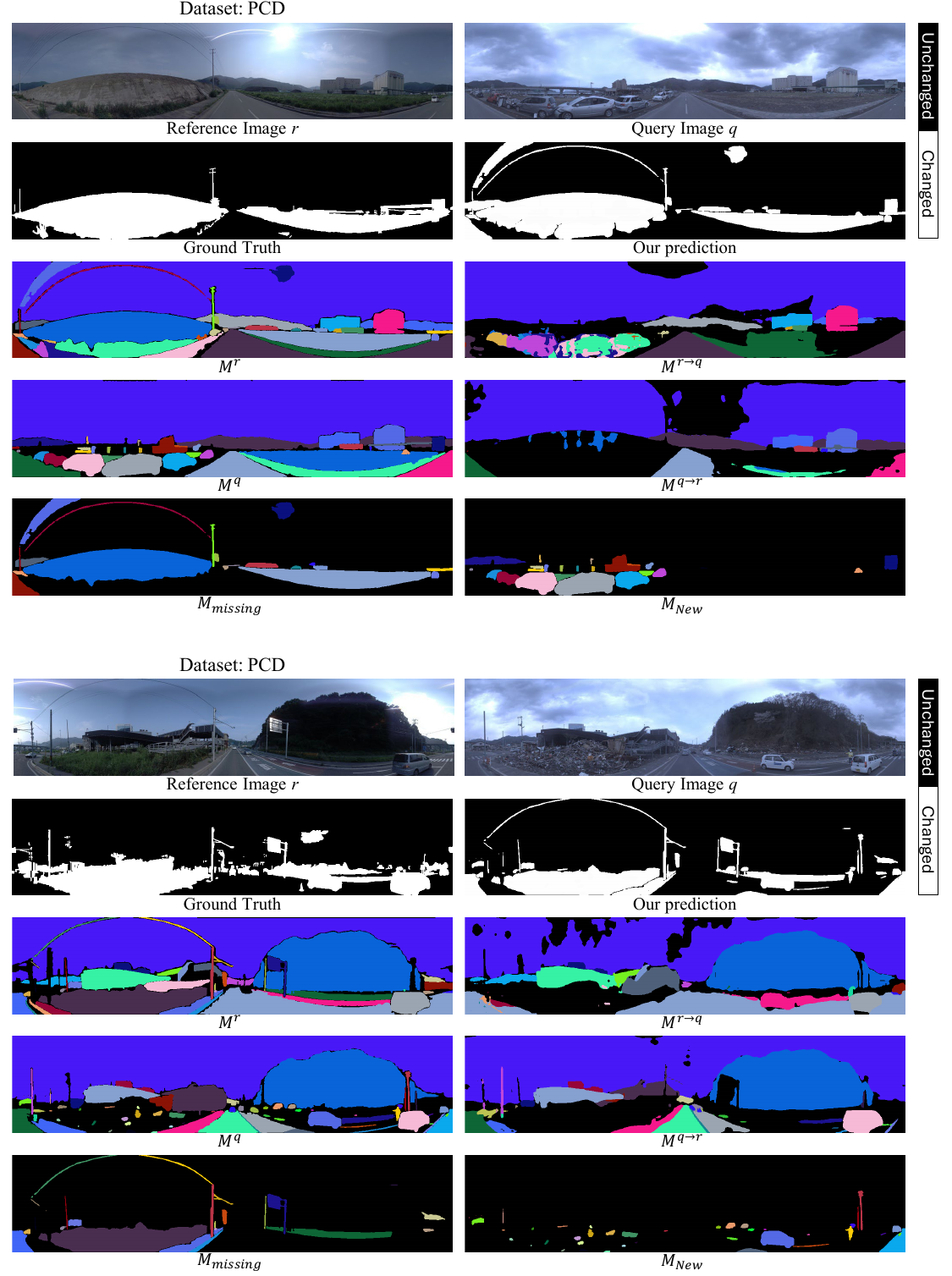}
  % \vspace{-0.5cm}
  \vspace{-0.2cm}
  \caption{\textbf{Qualitative results on PCD~\cite{jst2015change}.} 
 }
  \vspace{-0.2cm}
  \label{figure_qualitative4}
\end{figure*}

We present additional qualitative results in Figures~\ref{figure_qualitative1}, \ref{figure_qualitative2}, \ref{figure_qualitative3}, and \ref{figure_qualitative4} to show the effectiveness of our approach across various datasets. %Additionally, detailed images of the intermediate processes are included to help understand our approach.
To enhance understanding, detailed images of the intermediate processes are also provided.
During the intermediate process, identical masks before and after tracking are represented by the same color. Specifically, the same colored masks in $M^r$ and $M^{r \to q}$ denote the same object mask, and the same applies to $M^q$ and $M^{q \to r}$. However, since $M^r$ and $M^q$ do not share a tracking relationship, the same color between these two images holds no relationship.

The qualitative results show that our approach effectively identifies new and missing objects, and generates the final prediction accurately.

% \clearpage
% \bibliography{aaai25}
\end{document}